%% file: main.tex
\documentclass[runningheads]{llncs}

\usepackage{eccv}

\usepackage{eccvabbrv}

\usepackage{graphicx}
\usepackage{booktabs}

\usepackage[accsupp]{axessibility}  % Improves PDF readability for those with disabilities.

\usepackage{hyperref}

\usepackage{adjustbox}
\usepackage{multirow}
\usepackage{booktabs}
\usepackage{makecell}
\usepackage{pifont}
\usepackage{amssymb}
\usepackage{wrapfig}
\usepackage{bbding}

\usepackage{orcidlink}

\begin{document}

\title{OmniX: From Unified Panoramic Generation and Perception To Graphics-Ready 3D Scenes}

\titlerunning{OmniX}

\author{Yukun Huang\inst{1} \and
Jiwen Yu\inst{1,2} \and
Yanning Zhou\inst{3} \and
Jianan Wang\inst{4} \and
Xintao Wang\inst{2} \and
Pengfei Wan\inst{2} \and
Xihui Liu\inst{1}
}

\authorrunning{Huang et al.}

\institute{$^1$University of Hong Kong
\qquad
$^2$Kuaishou Technology\\
$^3$Tencent
\qquad
$^4$Astribot
}

% \maketitle

{
\maketitle
\begin{center}
\includegraphics[width=\linewidth]{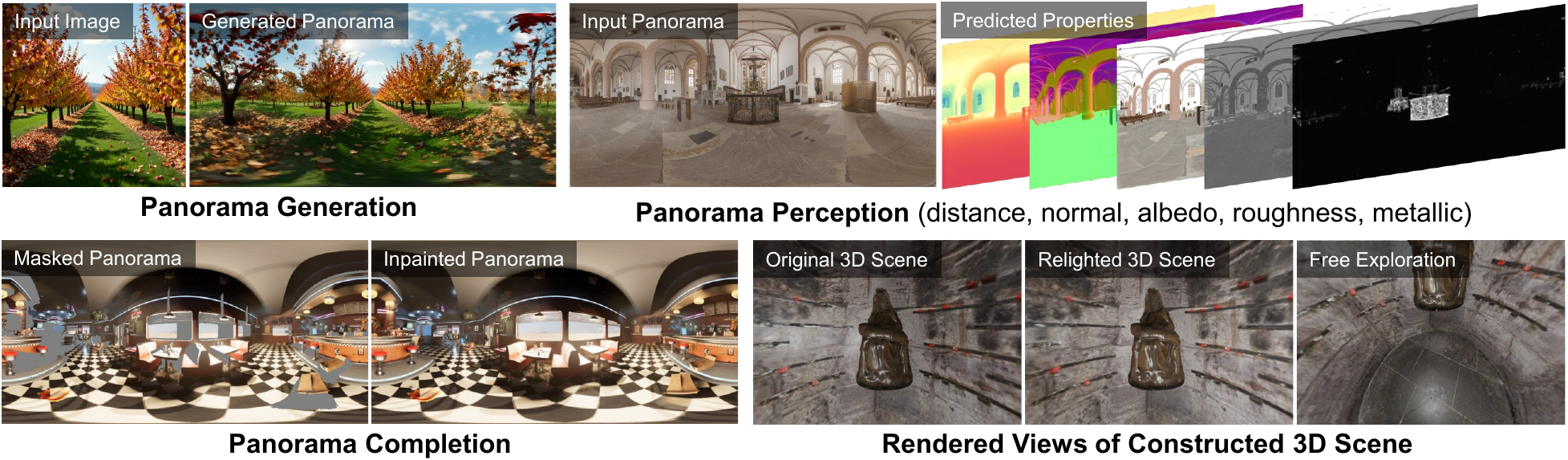}
\captionof{figure}{We present \textbf{OmniX}, a versatile and unified framework for panorama generation, perception, and completion. This framework enables the automatic construction of 3D scenes with support for physically based rendering (PBR) from input images.}
\label{fig:teaser}
\end{center}
}

\begin{abstract}
There are two prevalent ways for automatic 3D scene construction: procedural generation and 2D lifting. Among these, panorama-based 2D lifting has emerged as a promising technique, leveraging powerful 2D generative priors to produce immersive, realistic, and diverse 3D environments. In this work, we advance this technique to generate graphics-ready 3D scenes suitable for physically based rendering (PBR), relighting, and simulation. Our key insight is to repurpose 2D generative models for panorama perception of geometry, textures, and PBR materials. Unlike existing 2D lifting approaches that emphasize appearance generation and neglect the perception of intrinsic properties, we present OmniX, a versatile and unified framework for panorama generation, perception, and completion. Built upon cross-modal adapter structure and cyclic spatial operators, OmniX effectively repurposes pre-trained 2D flow matching priors for joint modeling of multimodal, seamless equirectangular representations. Furthermore, we construct a large-scale synthetic panorama dataset comprising high-quality multimodal panoramas from diverse indoor and outdoor scenes. Extensive experiments demonstrate the effectiveness and generality of OmniX as a unified framework for panorama generation and perception across geometry, lighting, and semantics, enabling graphics-ready 3D scene generation and opening new possibilities for immersive and physically realistic virtual world creation.
Project page is available at {\footnotesize\url{https://yukun-huang.github.io/OmniX/}}.
\end{abstract}

\input{sections/1_introduction}

\input{sections/2_related_work}

\input{sections/3_method}

\input{sections/4_experiment}

\input{sections/5_conclusion}

\section*{Acknowledgements}
The research work described in this paper was conducted in the JC STEM Lab of Autonomous Intelligent Systems funded by The Hong Kong Jockey Club Charities Trust.

\bibliographystyle{splncs04}
\bibliography{main}

\input{sections/X_suppl}

\end{document}

%% file: sections/1_introduction.tex
\section{Introduction}
\label{sec:intro}

Digitizing the 3D world we live in is a technological endeavor that is both imaginative and valuable. Digital replication~\cite{digital_cousin,litereality} of 3D scene allows us humans to obtain entertainment and interactive experiences that are difficult to obtain in daily life, or enables near-zero-cost simulation learning for intelligent agents or robots. However, constructing complex 3D scenes requires significant effort and time from artists and engineers, which limits the scale of 3D scene data and hinders the development of native 3D scene generative models.

To automatically build 3D scenes while circumventing data shortages, the community has leveraged large visual language foundation models trained on large-scale text, image, and video data. Based on these powerful models, two typical approaches emerge: procedural generation~\cite{infinitegen,layoutgpt} and 2D lifting~\cite{dreamfusion,wonderworld}. While procedural generation relies on retrieving objects from a 3D asset library to build the scene, 2D lifting methods directly repurpose 2D generative priors for 3D scene generation, achieving diverse and high-quality results. Recent works~\cite{layerpano3d,dreamcube,scenedreamer360,dreamscene360} further introduces panoramic representations, which serve as a bridge between 2D and 3D, greatly improving the cross-view consistency of generated 3D scenes. However, these works emphasize appearance generation rather than intrinsic perception, generally using off-the-shelf depth estimation models to extract scene geometry without textures and PBR materials. This hinders the integration of generated 3D scenes into modern graphics pipelines.

In this paper, we present \textbf{OmniX}, a versatile framework that repurposes pre-trained 2D flow matching models for panorama generation, perception, and completion. To this end, we introduce a unified formulation for panoramic vision tasks, extending the 2D generative paradigm to image-to-panorama generation, panorama-to-X perception, and their mask-guided generalization.

Specifically, to enhance multimodal panorama modeling, we introduce two key technical designs.
First, we develop \emph{circular synchronization} to mitigate the prevalent issue of discontinuous seams in equirectangular panorama generation. In contrast to prior approaches~\cite{diffusion360, panodiffusion} that directly manipulate latent features, our method provides a fundamental solution by enforcing circular translation equivariance in the model’s spatial operators. This structural mechanism ensures circular consistency without compromising the generative capacity of pre-trained models.
Second, we investigate cross-modal adapter architectures for handling diverse input modalities and introduce an effective and flexible design, \emph{modality-specific adapters}. These adapters integrate multiple LoRAs~\cite{lora}, allowing modality-specific adaptation while leveraging pre-trained 2D generative priors, thereby improving performance on cross-modal vision tasks.

In addition, we construct a synthetic panorama dataset, \textbf{PanoX}, covering indoor and outdoor scenes and various visual modalities such as distance, normal, albedo, roughness, and metallic. This dataset addresses the shortage of high-quality panorama data with dense geometry and material annotations.

Our main contributions are as follows:
\begin{itemize}

\item We present {OmniX}, a unified framework that repurposes pre-trained 2D flow matching models for panorama generation, perception, and completion.

\item To enhance multimodal panorama modeling, we introduce two key designs: {circular synchronization}, which ensures seamless equirectangular panoramas, and {modality-specific adapters}, which allow effective integration of diverse cross-modal inputs while leveraging pre-trained 2D generative priors.

\item We introduce {PanoX}, a synthetic panorama dataset covering both indoor and outdoor scenes. PanoX addresses the scarcity of high-quality panoramic data with dense geometric/material annotations.

\item Extensive experiments demonstrate the effectiveness of OmniX in panorama generation and perception. Moreover, our framework enables the automatic construction of immersive, PBR-ready 3D scenes from images.

\end{itemize}

%% file: sections/2_related_work.tex
\section{Related Work}
\label{sec:related_work}

\subsection{Inverse Rendering}
Inverse rendering~\cite{barrow1978recovering} aims to estimate intrinsic scene properties such as geometry, materials, and lighting from images. With the rapid progress of generative models, particularly diffusion models, researchers have explored their potential for inverse rendering~\cite{intrinsix,IDArb,diffusion_renderer,rgbx}. IntrinsiX~\cite{intrinsix} generates high-quality PBR maps from text prompts using a diffusion process, supporting precise material and lighting editing. DiffusionRenderer~\cite{diffusion_renderer} leverages video diffusion models for joint inverse and forward rendering, combining G-buffer estimation with photorealistic image generation through co-training on synthetic and real data.

Panoramic images capture a wider field of view and provide more comprehensive scene information, making them versatile for various applications. Yet, inverse rendering with panoramas remains underexplored. PhyIR~\cite{phyir} recovers geometry, complex SVBRDFs, and spatially-coherent illumination from a panoramic indoor image using an enhanced SVBRDF model and a physics-based in-network rendering layer to handle complex materials like glossy, metal, and mirror surfaces. 
However, it is limited to indoor scenes, while we leverage 2D generative priors to generalize across both indoor and outdoor environments.

\subsection{3D Scene Generation}

Existing 3D scene generation methods can be mainly divided into procedural generation~\cite{cityengine,musgrave1989synthesis,infinitegen,layoutgpt} and 2D lifting~\cite{dreamcube,immersegen,mvdiffusion,4real,wonderjourney}.

Procedural generation creates 3D scenes based on predefined rules or constraints. These methods are scalable and widely used in domains such as gaming, urban planning, and architecture, but often lack diversity and realism due to their rule-based nature. Representative works include CityEngine~\cite{cityengine}, which uses grammar-based rules for urban layouts, and InfiniGen~\cite{infinitegen}, which integrates terrain, material, and creature generators to produce diverse environments.

2D lifting methods bridge 2D inputs and 3D representations. Image-based approaches reconstruct 3D scenes from single or sequential images using outpainting or depth estimation, with works like ImmerseGAN~\cite{immersegen} and MVDiffusion~\cite{mvdiffusion} generating panoramas for scene synthesis. Video-based methods leverage temporal information to ensure coherent dynamic scenes, exemplified by VividDream~\cite{vividdream} and 4Real~\cite{4real}. These methods emphasize appearance generation, relying on off-the-shelf depth estimators for geometry while neglecting intrinsic properties such as albedos, normals, and PBR materials.

%% file: sections/3_method.tex
\section{Method}
\label{sec:method}

\subsection{Overview}
We first construct a multimodal synthetic panorama dataset, PanoX (Sec.~\ref{sec:method_data}), which provides high-quality supervision for panorama understanding. Building upon this dataset, we introduce OmniX, a versatile and unified framework that repurposes pre-trained 2D flow matching models~\cite{sd3,flow_matching} for panorama generation, perception, and completion (Sec.~\ref{sec:method_omnix}). Finally, We demonstrate OmniX's potential in PBR-ready 3D scene generation (Sec.~\ref{sec:method_scene}).

\subsection{PanoX: A Synthetic Panorama Dataset with Dense Annotations}
\label{sec:method_data}

Omnidirectional visual perception is crucial for visual understanding and spatial intelligence. To effectively learn perception across a wide field of view (FoV), large-scale panorama datasets with dense annotations are necessary. While several narrow FoV image datasets~\cite{hypersim,interiorverse,matrixcity} offer rich geometry and material annotations, there remains a scarcity of panorama datasets equipped with dense annotations within the research community.

\begin{figure}[b]
\centering
\includegraphics[width=\linewidth]{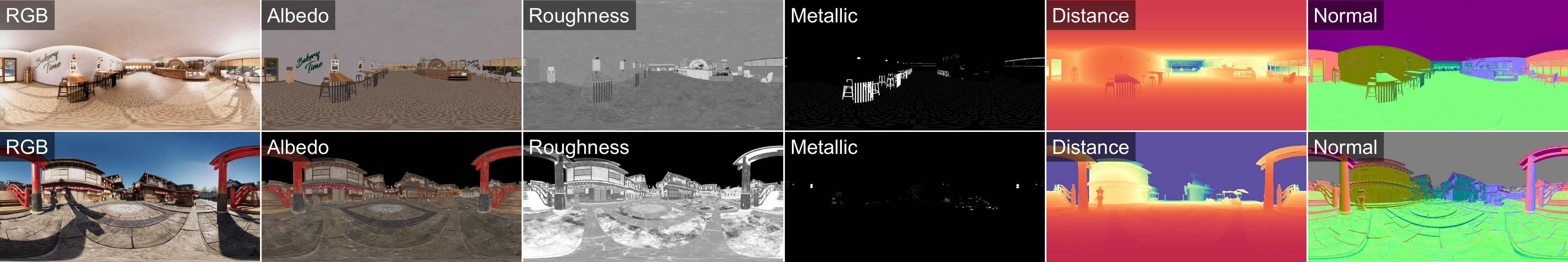}
\caption{\textbf{A preview of the proposed PanoX dataset}, providing high-quality panoramic rendered images with rich pixel-aligned annotations, including albedo, roughness, metallic, distance, and normal, across both indoor and outdoor scenes.}
\label{fig:panox}
\end{figure}

To this end, we introduce PanoX, a multimodal synthetic panorama dataset with dense geometry and material annotations. Given the challenges of collecting real-world panorama data and the high cost of manual annotation, we leverage synthetic 3D scene assets and Unreal Engine 5 to generate pixel-aligned multimodal panorama data. A preview of PanoX is shown in Figure~\ref{fig:panox}.

Specifically, PanoX comprises eight large-scale 3D environments, including five indoor and three outdoor scenes such as stores, warehouses, and wilderness areas. Each scene is rendered into RGB panoramas, along with corresponding distance, world normal, albedo, roughness, and metallic. We also provide text descriptions corresponding to the panoramic images, which are extracted using Florence 2~\cite{florence2}. The entire dataset contains more than 10,000 instances, corresponding to 60,000 panoramic images of different modalities. We split the samples from the first six scenes into training, validation, and test segments in an 8:1:1 ratio, obtaining PanoX-Train, PanoX-Val, PanoX-Test. The remaining two scenes are grouped as out-of-domain test set (\ie, PanoX-OutDomain) for generalization evaluation.

We compare the proposed PanoX dataset with existing image datasets rendered from synthetic 3D scenes, as shown in Table~\ref{tab:comp_dataset}. To the best of our knowledge, the proposed PanoX is the first panorama dataset covering both indoor and outdoor scenes with dense geometry and material annotations.

\input{tables/comp_dataset}

\subsection{OmniX: A Unified Framework for Panorama Generation, Perception, and Completion}
\label{sec:method_omnix}

OmniX is a versatile framework for unified panorama perception and generation, built on the pre-trained 2D flow matching model~\cite{flux1dev}. To this end, we provide a general formulation for unified generation and perception.

\begin{figure}[t]
\centering
\includegraphics[width=\linewidth]{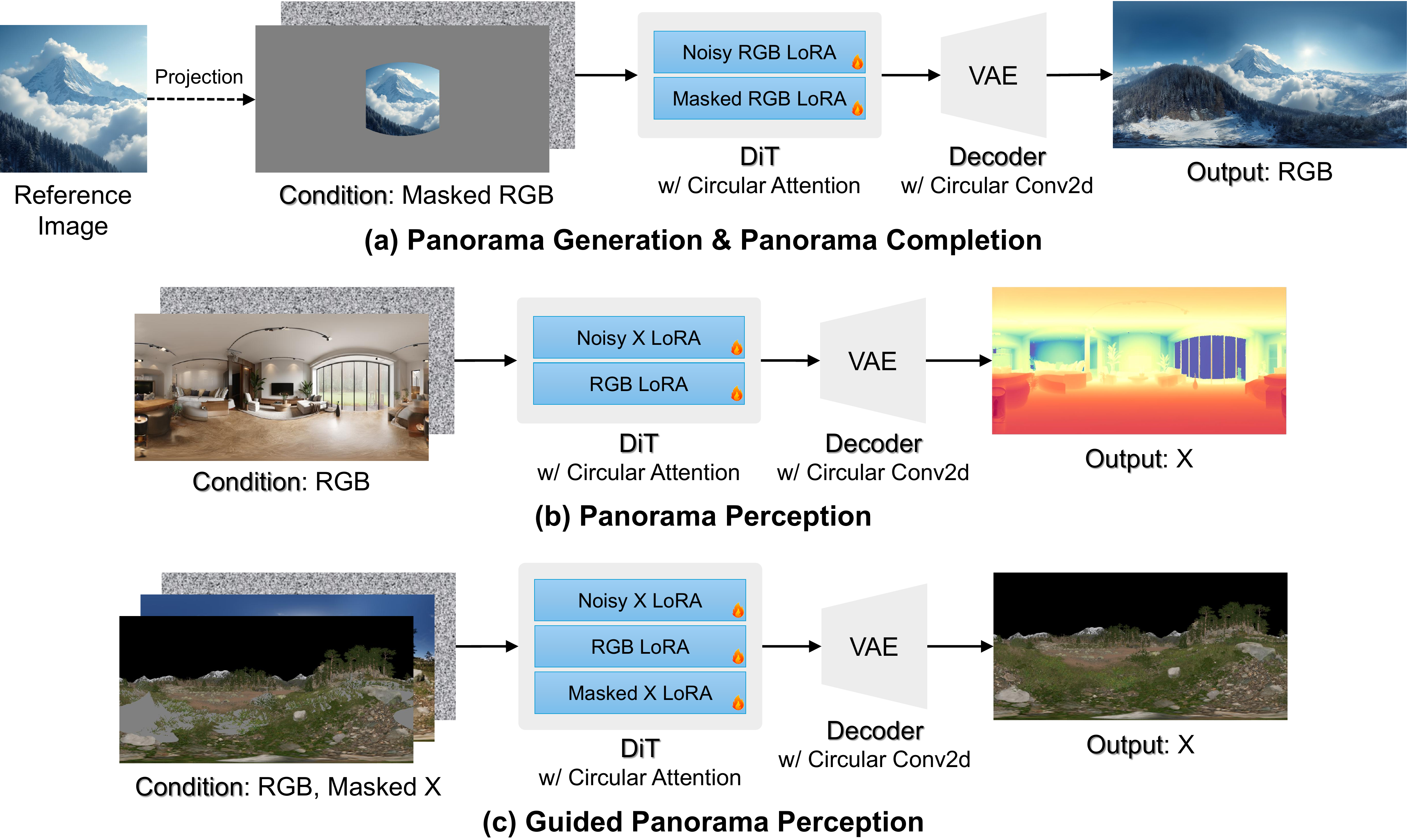}
\caption{\textbf{OmniX} is a versatile and unified framework that repurposes pre-trained 2D flow-matching models for panorama generation, perception, and completion. Built upon circular synchronization operators and modality-specific LoRAs, OmniX flexibly adapts to different panorama tasks and produces seamless results.}
\label{fig:method_framework}
\end{figure}

\textbf{Unified formulation.} Typically, a flow matching-based image generator $f_{\theta}$ is trained to predict the velocity vector $\mathbf{v}$ from latent representation $\mathbf{z}_0$ to latent representation $\mathbf{z}_1$, given a textual prompt $y$ and the current timestep $t$:
\begin{equation}
\mathbf{v}_{t} = f_{\theta}(\mathbf{z}_t, y, t),
\end{equation}
The predicted target $\hat{z}_1$ can be obtained by solving the following ordinary differential equation (ODE):
\begin{equation}
\hat{\mathbf{z}}_1 = \mathbf{z}_0 + \int_0^1\mathbf{v}_t~dt = \mathbf{z}_0 + \int_0^1f_\theta(\mathbf{z}_t, y, t) dt.
\end{equation}
Our goal is to expand this image generation paradigm into a unified panorama generation, perception, and completion framework. To this end, we generalize the model $f_{\theta}$ to take multiple condition inputs:
\begin{equation}
\hat{\mathbf{z}}_1 = \mathbf{z}_0 + \int_0^1f_\theta(\mathbf{z}_t, \mathbf{c}^0, \mathbf{c}^1, ..., y, t) dt,
\end{equation}
where $\{\mathbf{c}^i | i=0, 1, ...\}$ denotes the conditioning inputs, each spatially aligned with $\mathbf{z}_t$. The modality and number of the conditioning inputs depend on the target task. Figure~\ref{fig:method_framework} shows three representative target tasks, each corresponding to a different configuration of conditioning inputs. Detailed descriptions of these task settings are provided in the supplementary material.
% For example, $\mathbf{c}^i$ may correspond to RGB images for RGB-to-X generation. Further discussion is provided in the supplementary material.

\textbf{Circular synchronization for seamless generation.} Circular blending~\cite{diffusion360} is a commonly used technique~\cite{layerpano3d,hunyuan3dworld} for achieving seamless panorama generation with latent diffusion models~\cite{latentdiffusion}, which weights and blends the leftmost and rightmost edges of latent features to ensure closed loops. While effective, such latent-level operations may introduce unwanted feature distortions near the boundary. We argue that seam discontinuities fundamentally originate from the lack of circular translation equivariance in spatial operators (\eg, attentions and convolutions). To address this issue, we propose Circular Synchronization, a training-free approach that adapts spatial operators for seamless generation without fine-tuning the pre-trained model parameters.

\begin{figure}[t]
\centering
\includegraphics[width=\linewidth]{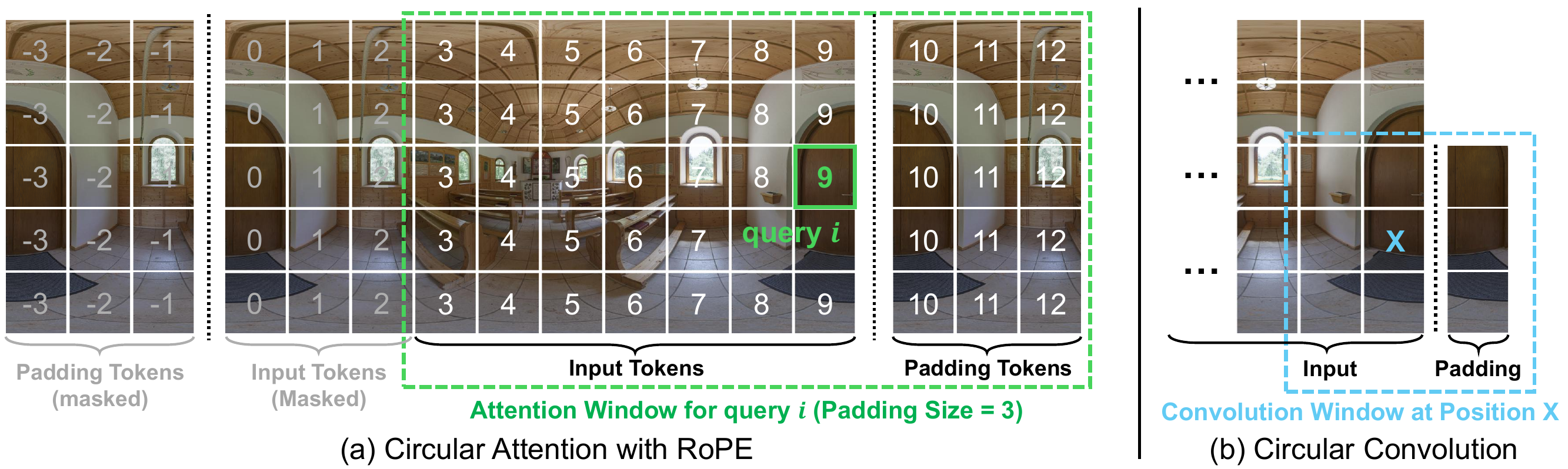}
\caption{\textbf{Circular Synchronization} adapts the receptive window of spatial operators to enforce circular translational equivariance, including: (a) Circular Attention with RoPE~\cite{rope}, where the numbers in the image patches indicate horizontal position IDs, and (b) Circular Convolution (also referred to as circular padding).}
\label{fig:method_circular}
\end{figure}

The core idea of circular synchronization is to adapt all spatial operators to be circular translation-equivariant, as illustrated in Figure~\ref{fig:method_circular}. For convolutions, this adaptation is straightforward, as it only requires replacing the conventional padding strategy with circular padding~\cite{360dvd}. For attentions equipped with RoPE~\cite{rope}, the adaptation involves two steps: token padding and attention masking. Specifically, token padding applies circular padding to the left and right boundaries of key and value tokens (excluding query tokens), with the padding size controlled by a padding-size hyperparameter. Attention masking constrains the attention window to ensure a circular and spatially uniform receptive field for each query token.
By enforcing circular synchronization across all convolutional and attention operations in the pre-trained DiT~\cite{dit} and VAE, our method enables seamless circular image generation, as shown in Figure~\ref{fig:comp_circular}.

\begin{figure}[t]
\centering
\includegraphics[width=\linewidth]{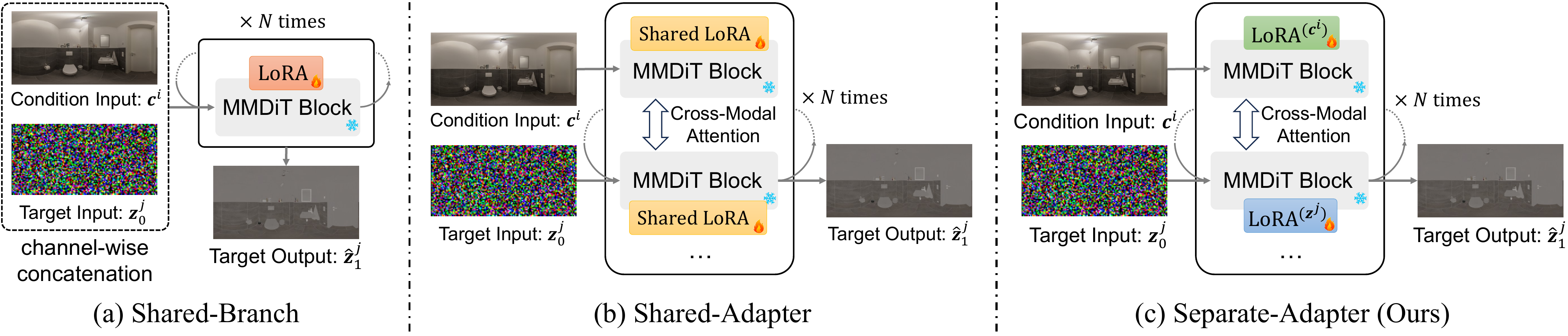}
\caption{\textbf{Different cross-modal adapter structures} for multiple condition inputs $\{\mathbf{c}^i~|~i=0, 1, ...\}$ and multiple target outputs $\{\mathbf{\hat{z}}_1^j~|~j=0, 1, ...\}$. Specifically, (a) \emph{Shared-Branch} concatenates different inputs along the channel dimension; (b) \emph{Shared-Adapter} performs token-wise concatenation across multiple inputs; and (c) \emph{Separate-Adapter} builds upon \emph{Shared-Adapter} by assigning modality-specific weights to each input type.}
\label{fig:method_adapter}
\end{figure}

For multi-view methods~\cite{cubediff,recipe,mvdiffusion}, boundary consistency is typically achieved through overlapping fields of view (FoVs), where adjacent views provide mutual constraints for alignment. Such strategies require additional view aggregation mechanisms and computational overhead, while treating panoramic consistency as an external constraint. In contrast, our Circular Synchronization directly adapts spatial operators to be circular translation-equivariant, making boundary consistency an intrinsic property of the generation process. This design preserves the pretrained generative priors without modifying the underlying model parameters and produces more coherent boundary transitions.

\textbf{Modality-specific adapters for cross-modal generation.} We explore multiple ways to adapt the pre-trained DiT for cross-modal 2D inputs, as shown in Figure~\ref{fig:method_adapter}. Specifically, depending on how branches and adapters are shared, these methods can be divided into: Shared-Branch, Shared-Adapter, and Separate-Adapter. Given multiple 2D inputs, Shared-Branch and Shared-Adapter perform channel-wise and token-wise concatenations, respectively. Building upon token-wise concatenation, Separate-Adapter further assigns different LoRAs~\cite{lora} to different types of inputs. Note that all 2D inputs and outputs are spatially aligned and share the same 2D positional encoding.

Empirical results indicate that the Separate-Adapter design not only achieves superior performance in cross-modal panorama perception tasks (as reported in Table~\ref{tab:ablation_adapter}) but also allows flexible expansion to new input modalities with minimal impact on the model’s weight distribution. Therefore, OmniX employs this modality-specific adapter design.

\textbf{Optimization.} Built upon modality-specific adapters, multiple LoRAs~\cite{lora} are trained to exploit a pre-trained 2D flow matching model for feature extraction from conditional inputs and predicting velocity vectors for target outputs. While both the condition $\mathbf{c}$ and the target $\mathbf{z}_t$ are input to DiT, only the target output is used to compute the flow matching loss~\cite{flow_matching}:
\begin{equation}
\mathcal{L}=\mathbb{E}_{t,\mathbf{z}_1,\mathbf{z}_0}\|\mathbf{v} - f_\theta(\mathbf{z}_t,\mathbf{c},t)\|^2,
\end{equation}
where the velocity vector $\mathbf{v}=\mathbf{z}_1-\mathbf{z}_0$. Note that this objective can be generalized to \textbf{M}ultiple condition \textbf{I}nputs $\{\mathbf{c}^i~|~i=0, 1, ...\}$ and \textbf{M}ultiple target \textbf{O}utputs $\{\mathbf{z}_1^j~|~j=0, 1, ...\}$, yielding a MIMO version of flow matching loss:
\begin{equation}
\mathcal{L}_\text{mimo}=\mathbb{E}_{t,\mathbf{z}_1^j,\mathbf{z}_0}\|\mathbf{v} - f_\theta(\mathbf{z}_t,\mathbf{c}^0,\mathbf{c}^1,...,t)\|^2.
\end{equation}
Note that circular synchronization is a structural constraint and does not need to be performed during training, which avoids additional computational costs.

\subsection{Application: PBR-Ready 3D Scene Generation}
\label{sec:method_scene}
The OmniX framework opens up the possibility of automatically constructing PBR-ready 3D scenes from images or panoramas. Specifically, the entire pipeline consists of three stages: (a) multimodal panorama generation, (b) scene reconstruction, and (c) iterative completion.

\textbf{Multimodal panorama generation.} OmniX offers a general solution for image-to-panorama generation and RGB-to-X panorama perception. We train multiple adapters to repurpose the pre-trained flow matching model for these tasks. By switching adapters of different tasks, we can achieve a generative chain of ``image $\rightarrow$ panorama $\rightarrow$ panorama with intrinsic properties''.

\textbf{Scene reconstruction.} Given a panoramic distance map, since the ray direction corresponding to each pixel is known, the pixels can be projected into 3D space as vertices of a 3D mesh. The connectivity of these vertices can be further determined based on pixel neighbors and relative distances. Once the 3D mesh of the scene is obtained, the panoramic maps of other modalities (\ie, albedo, normal, roughness, and metallic) can be assigned to each triangle face via spherical UV unwrapping, resulting in a PBR-ready scene-level 3D asset.

\begin{figure}[t]
\includegraphics[width=\linewidth]{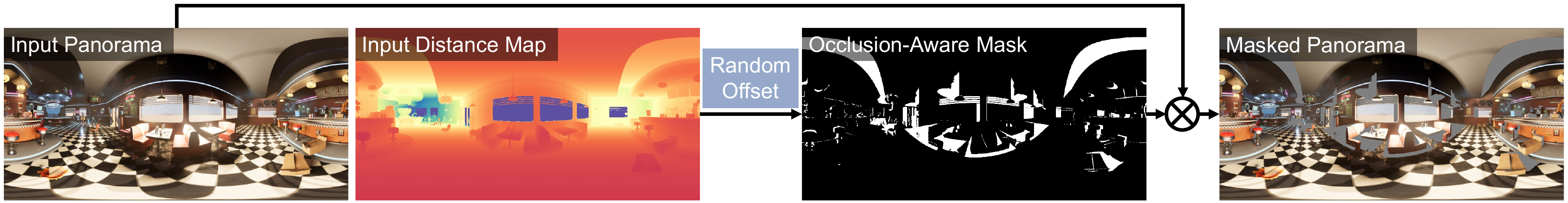}
\caption{\textbf{Occlusion-aware masked data construction pipeline.} Based on the panoramic distance map and a randomly sampled 3D displacement, we can estimate the occluded regions by ray intersection. These regions are used as masks to construct masked data for training the panorama completion models.}
\label{fig:method_mask}
\end{figure}

\textbf{Iterative completion.} A panoramic view provides only an omnidirectional observation from a fixed viewpoint, and thus the reconstructed scene does not support free exploration. Iterative scene completion~\cite{layerpano3d,wonderworld} is therefore essential for constructing explorable and potentially large-scale 3D scenes. To this end, we augment the OmniX adapters with mask inputs and fine-tune them for completion and guided perception, resulting in OmniX-Fill. Specifically, to simulate occlusion-induced scene holes, we propose a depth-based sampling strategy to generate occlusion-aware masks, as illustrated in Figure~\ref{fig:method_mask}. Through the interaction between OmniX-Fill and the graphics engine, we are able to generate new regions while preserving previously reconstructed content.

%% file: tables/comp_dataset.tex
\begin{table}[t]
\begin{center}
\caption{Comparison between PanoX and existing synthetic scene datasets.}
\label{tab:comp_dataset}
\footnotesize
\renewcommand{\arraystretch}{0.95}
\setlength{\tabcolsep}{7pt}
\resizebox{\linewidth}{!}{
\begin{tabular}{lccccc}
    \toprule
    \textbf{Dataset} & Geometry & Material & \makecell{Include Panorama} & Indoor & Outdoor \\
    \midrule
    InteriorNet~\cite{interiornet} & \checkmark & \checkmark & \checkmark$^{\dagger}$ & \checkmark & \ding{53} \\ 
    Structured3D~\cite{structured3d} & \checkmark & \ding{53} & \checkmark & \checkmark & \ding{53} \\ 
    Hypersim~\cite{hypersim} & \checkmark & \ding{53} & \ding{53} & \checkmark & \ding{53} \\ 
    InteriorVerse~\cite{interiorverse} & \checkmark & \checkmark & \ding{53} & \checkmark & \ding{53} \\ 
    FutureHouse$^{\ddagger}$~\cite{phyir} & \checkmark & \checkmark & \checkmark & \checkmark & \ding{53} \\ 
    MatrixCity~\cite{matrixcity} & \checkmark & \checkmark & \ding{53} & \ding{53} & \checkmark \\ 
     PanoX (Ours) & \checkmark & \checkmark & \checkmark & \checkmark & \checkmark \\ 
    \bottomrule
\end{tabular}
}
\end{center}
\footnotesize{$^\dagger$ InteriorNet only contains panoramic RGB images without dense annotations.}\\
\footnotesize{$^{\ddagger}$ FutureHouse is no longer publicly available.}
\end{table}

%% file: sections/4_experiment.tex
\section{Experiment}
\label{sec:experiment}

\subsection{Implementation Details}
Our method is implemented in PyTorch, trained and evaluated on four NVIDIA L40S GPUs. We use the same optimization settings for all tasks. Specifically, we adopt an AdamW optimizer with learning rate of 1e-4 for training. No learning rate decay strategy is employed.
All panoramic images are resized to a resolution of $512\times1024$ for training, using the batch size of 1 for each GPU.

We trained 12 adapter models based on Flux.1-dev~\cite{flux1dev} for different panoramic vision tasks, including: image-to-panorama generation, panorama perception (distance, normal, albedo, roughness, metallic), and their corresponding masked versions. Each of the adapter model consists of two or more LoRAs, depending on the number of input modalities. 

\subsection{Results on Panorama Generation}
For panorama generation, we consider a perspective-to-panorama setting~\cite{cubediff}, in which text–image pairs are used as conditioning inputs.

\textbf{Datasets.} For training, we combine multiple panorama datasets and publicly available sources, including Pano360~\cite{pano360}, Structured3D~\cite{structured3d}, Poly-Haven~\cite{polyhaven}, Humus~\cite{humus}, and our proposed PanoX. For evaluation, we report results on the Laval Indoor~\cite{laval} and SUN360~\cite{sun360} datasets following CubeDiff~\cite{cubediff}. The text prompts for all panoramic images are extracted by Florence 2~\cite{florence2}.

\textbf{Evaluation metrics.} To evaluate visual quality, we follow CubeDiff~\cite{cubediff} and report FID~\cite{fid}, KID~\cite{kid}, CLIP-FID~\cite{clip_fid}, and FAED~\cite{panfusion}. FID, KID, and CLIP-FID are computed on perspective crops from generated ERP panoramas, while FAED is evaluated directly on ERP outputs. We additionally report CLIP Score (CS)~\cite{clip_score} to measure text alignment.

\input{tables/comp_appearance}

\begin{figure}[t]
\centering
\includegraphics[width=\linewidth]{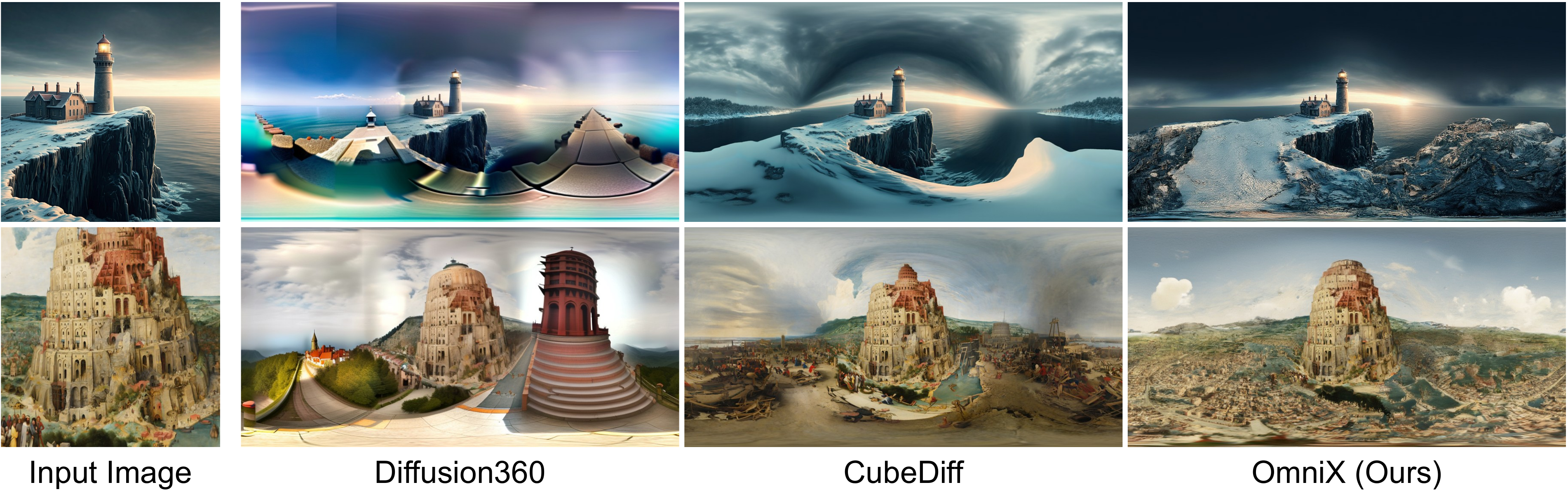}
\caption{\textbf{Qualitative evaluation of OmniX on panorama generation} compared to two image-to-panorama methods: Diffusion360~\cite{diffusion360} and CubeDiff~\cite{cubediff}.}
\label{fig:comp_appearance}
\end{figure}

\textbf{Quantitative evaluation.} As shown in Table~\ref{tab:comp_appearance}, OmniX achieves state-of-the-art performance across both the Laval Indoor and SUN360 datasets. On Laval Indoor, it obtains the best results on FID (7.4), CLIP-FID (2.5), FAED (5.2), and CLIP Score (28.47), while remaining competitive on KID, outperforming previous methods in both visual fidelity and text alignment. On SUN360, OmniX continues to deliver strong results, achieving the best KID (0.66), CLIP-FID (6.5), and CLIP Score (27.54), while maintaining competitive performance on FID and FAED. Compared to strong baselines like CubeDiff, OmniX demonstrates more consistent improvements across metrics, highlighting its ability to produce visually realistic and semantically aligned panoramic images.

\textbf{Qualitative evaluation.} As illustrated in Figure~\ref{fig:comp_appearance}, OmniX outperforms previous perspective image-to-panorama methods, including Diffusion360~\cite{diffusion360} and CubeDiff~\cite{cubediff}, in terms of panorama quality and scene consistency. Diffusion360 suffers from noticeable visual artifacts and inconsistent scene layouts, while CubeDiff alleviates some appearance artifacts but still produces structurally inconsistent panoramas, particularly for large-scale scene layouts requiring long-range spatial reasoning. In contrast, OmniX generates panoramas with improved structural fidelity, seamless global consistency, and realistic scene composition, while preserving fine-grained local details and coherent geometry across the entire 360° field of view.

\subsection{Results on Panorama Perception}
We divide panorama perception into intrinsic decomposition (albedo, roughness, metallic) and geometry estimation (distance, normal), and present both qualitative and quantitative results compared to competing methods.

\input{tables/comp_intrinsic}

\textbf{Datasets.} We use the standard splits of both PanoX and Structured3D for training and evaluation. During training, each batch is sampled from these two data sources with equal probability. Since Structured3D does not include PBR materials, only PanoX is used for roughness and metallic models.

\textbf{Evaluation metrics.} For visual perception tasks with ground truths, we adopt a variety of quantitative metrics for different modalities. Specifically, we use PSNR (Peak Signal-to-Noise Ratio) and LPIPS~\cite{lpips} as metrics for albedo, roughness, and metallic. For Euclidean distances, we use four commonly used metrics: AbsRel, $\delta$-1.25, MAE, and RMSE, following the implementation in~\cite{cheng2018depth}. For surface normals, We measure the pixel-wise angular error with ground truth and report the mean, median, and the percentage of pixels with an error below 5$^\circ$ and 30$^\circ$ following~\cite{rethinking_normal}. Note that there may be invalid values in the ground truths (\eg, pixels at infinite distance), we exclude these invalid values when calculating the metrics.

\begin{figure}[t]
\centering
\includegraphics[width=\linewidth]{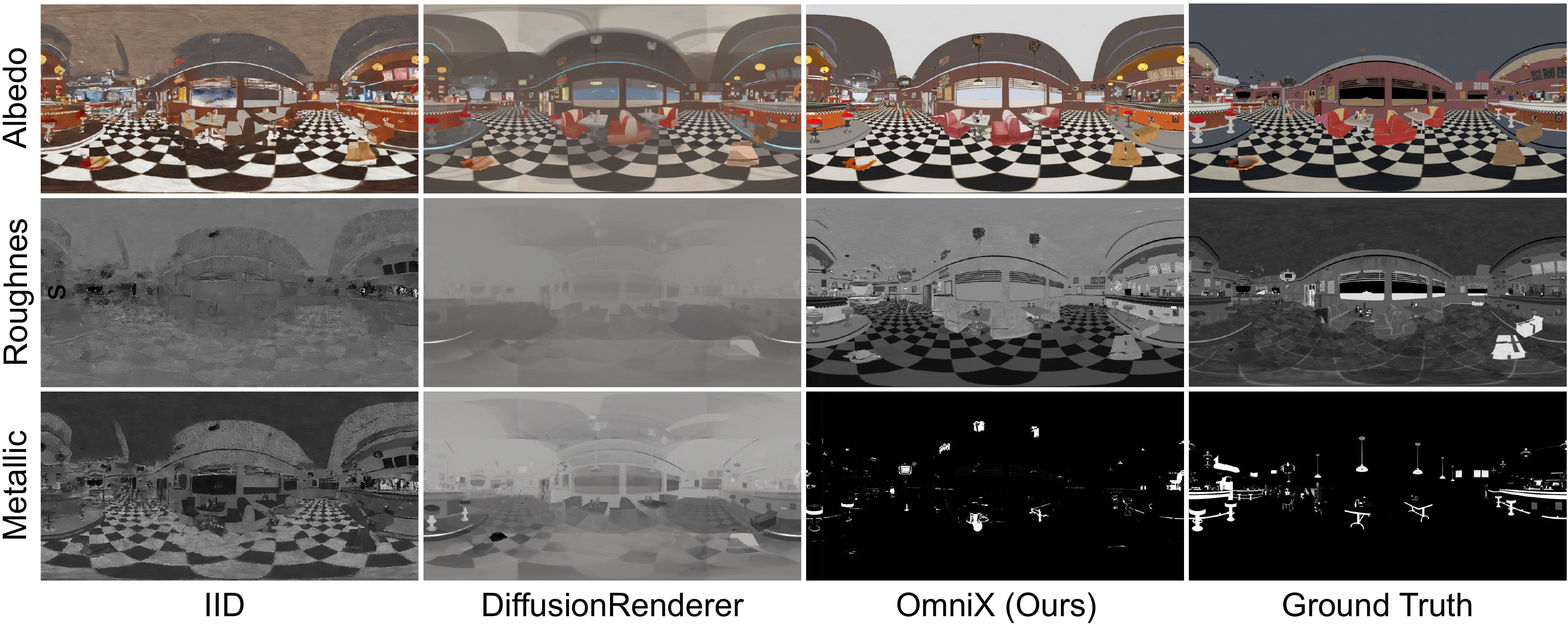}
\caption{\textbf{Qualitative evaluation of OmniX on panoramic intrinsic estimation} compared to two competing methods: IID~\cite{intrinsic_image_diffusion} and DiffusionRenderer~\cite{diffusion_renderer}. }
\label{fig:comp_intrinsic}
\end{figure}

\textbf{Panoramic intrinsic decomposition.} We compare our OmniX with five state-of-the-art intrinsic decomposition methods: RGB$\leftrightarrow$X~\cite{rgbx}, MGNet~\cite{interiorverse}, IDArb~\cite{IDArb}, IID~\cite{intrinsic_image_diffusion}, DiffusionRenderer~\cite{diffusion_renderer}. Note that DiffusionRenderer is a video-based inverse rendering method, so we render each panorama into multiple frames to fit its input. The quantitative results are reported in Table~\ref{tab:comp_intrinsic}. Our method achieves consistent state-of-the-art performance on the prediction of three intrinsic properties: albedo, roughness, and metallic. A qualitative comparison is shown in Figure~\ref{fig:comp_intrinsic}  to illustrate the prediction results.

\begin{figure}[t]
\centering
\includegraphics[width=\linewidth]{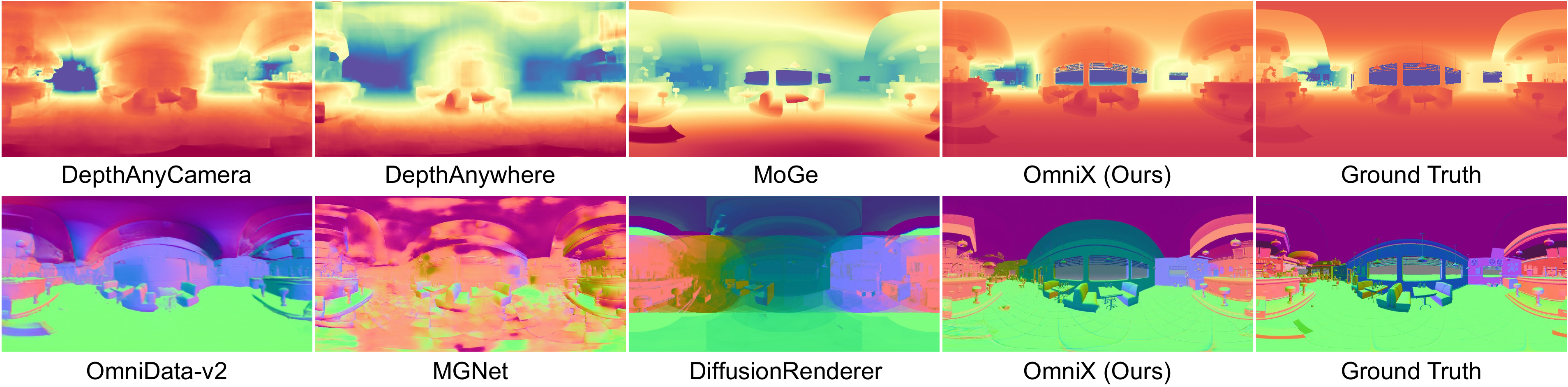}
\caption{\textbf{Qualitative evaluation of OmniX on panoramic geometry estimation} compared to state-of-the-art geometry estimation methods: DepthAnyCamera~\cite{depthanycamera}, DepthAnywhere~\cite{depthanywhere}, OmniData-v2~\cite{omnidata_v2}, MGNet~\cite{interiorverse}, DiffusionRenderer~\cite{diffusion_renderer}, and MoGe~\cite{moge}. Our method shows a notable advantage in capturing fine image details.}
\label{fig:comp_geometry}
\end{figure}

\textbf{Panoramic geometry estimation.} We compare our OmniX with two panoramic geometry estimation methods: DepthAnyCamera~\cite{depthanycamera}, DepthAnywhere~\cite{depthanywhere}, and four narrow-FoV geometry estimation methods: 
OmniData-v2~\cite{omnidata_v2}, MGNet~\cite{interiorverse}, DiffusionRenderer~\cite{diffusion_renderer}, and MoGe~\cite{moge}. The quantitative results are reported in Table~\ref{tab:comp_geometry}, where we achieve the highest normal estimation accuracy and and the second highest depth estimation accuracy. Note that MoGe~\cite{moge} integrate 21 large-scale datasets for training, while we use much less data to achieve competitive performance. We further provide a qualitative comparison in Figure~\ref{fig:comp_geometry} to illustrate the prediction results.

\input{tables/comp_geometry}

\textbf{In-the-wild panorama perception.} We empirically find that the proposed OmniX demonstrates superior generalization performance and is able to achieve satisfactory prediction results on in-the-wild images from the Internet. These results are presented in the supplementary material.

\subsection{Ablation Analysis and Discussion}
We conduct ablation studies to evaluate the key components of OmniX, with additional analysis and discussion provided in the supplementary material.

\input{tables/ablation_adapter}

\textbf{Cross-modal adapter structures.} We investigate the effect of different adapter structures on panorama perception performance, as reported in Table~\ref{tab:ablation_adapter}. Among the evaluated variants, the Separate-Adapter design adopted in OmniX achieves the best performance. This improvement stems from its ability to enable modality-specific adaptation while effectively leveraging pre-trained 2D generative priors. By decoupling adaptations across modalities without disrupting the original weight distribution, the proposed structure preserves the strengths of the pre-trained model while enhancing cross-modal learning capacity.

\textbf{Leveraging 2D generative priors.} We compare training from scratch with initialization from 2D generative pre-trained weights to evaluate their impact on panorama perception performance, as shown in Table~\ref{tab:ablation_adapter}. Initializing with 2D generative priors leads to consistent and substantial performance improvements, even for modalities with significantly different pixel-value distributions. These results highlight the strong transferability and effectiveness of 2D generative priors for panorama perception tasks.

\begin{figure}[t]
\centering
\includegraphics[width=\linewidth]{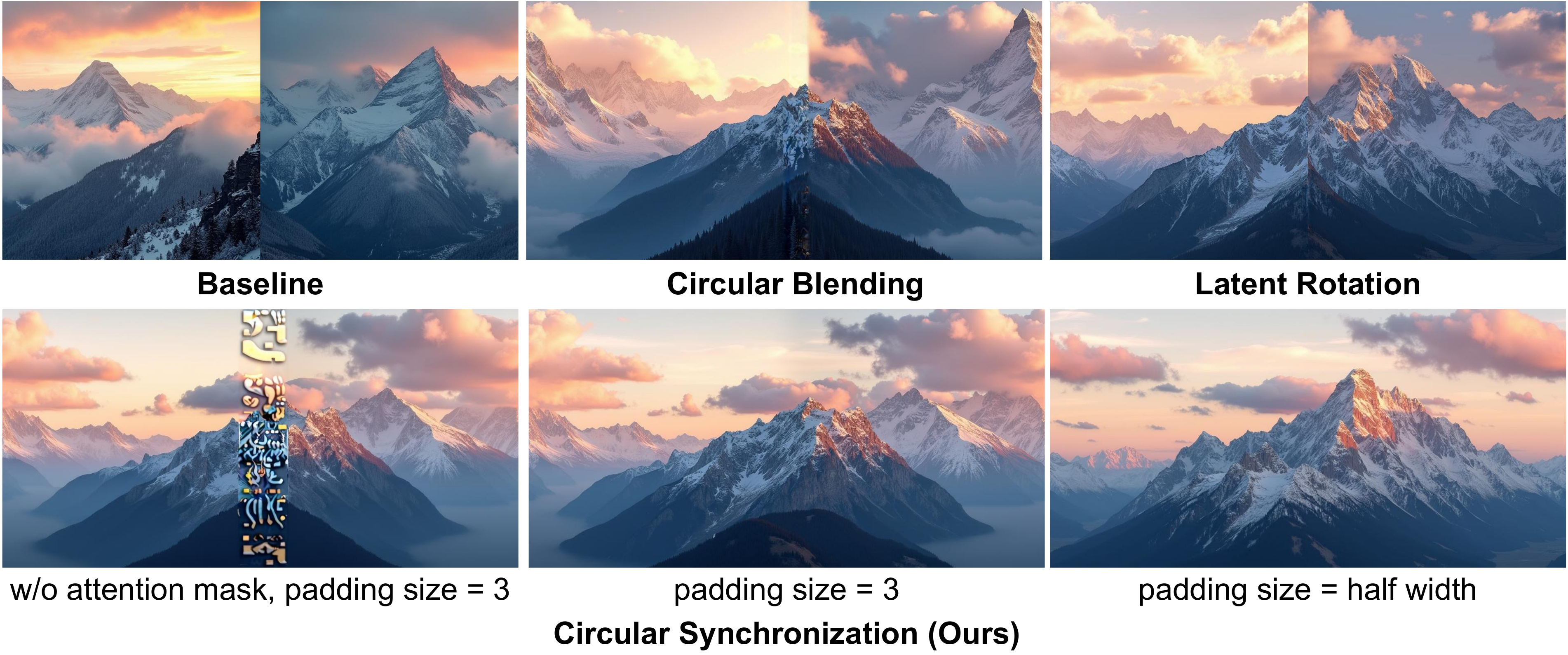}
\caption{\textbf{Qualitative analysis of circular synchronization.} The displayed images are generated by Flux.1-dev~\cite{flux1dev} and horizontally rolled by half their width to show the continuity at the seams. Compared to circular blending~\cite{diffusion360} and latent rotation~\cite{panodiffusion}, our method achieves near-perfect seamless image generation without any fine-tuning.}
\label{fig:comp_circular}
\end{figure}

\textbf{Circular synchronization.} We compare circular synchronization with baseline (Flux.1-dev~\cite{flux1dev}), circular blending~\cite{diffusion360}, and latent rotation~\cite{panodiffusion}, as shown in Figure~\ref{fig:comp_circular}. To better visualize seam continuity, the generated images are horizontally rolled by half of their width. The baseline exhibits clear discontinuities at the boundary, while circular blending and latent rotation partially alleviate seam artifacts but still introduce visible structural inconsistencies or texture distortions near the boundary. In contrast, circular synchronization produces nearly seamless transitions across the horizontal wrap-around. This result demonstrates that enforcing circular translation equivariance at the structural level effectively guarantees global consistency along the seam while preserving the generative fidelity of pre-trained 2D flow matching models.

\subsection{Applications}

\begin{figure}[tbp]
\centering
\includegraphics[width=\linewidth]{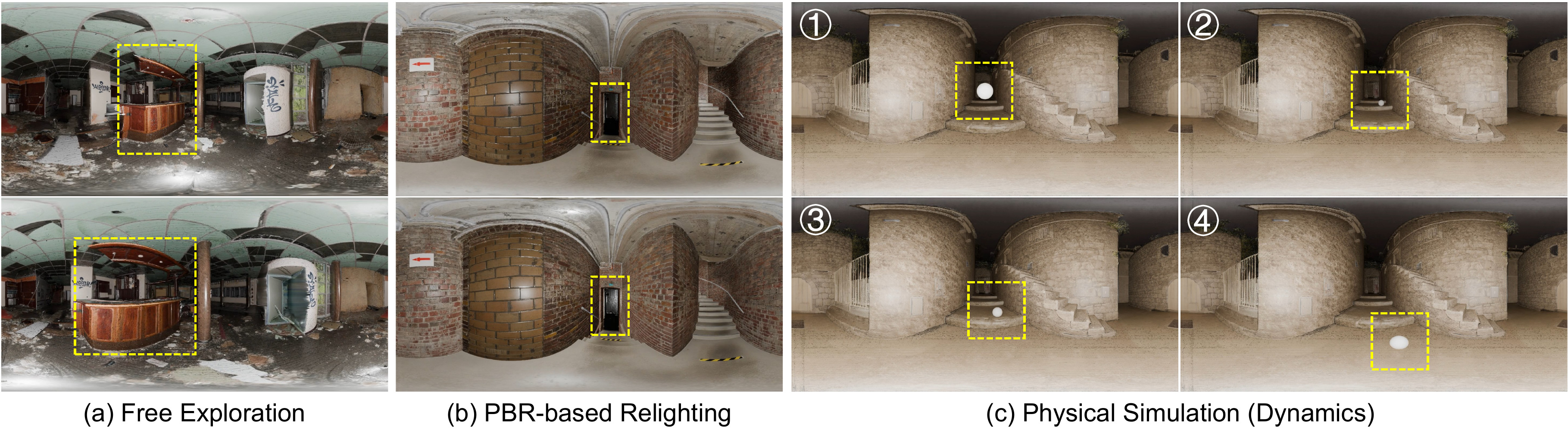}
\caption{\textbf{Demonstrations of the PBR-ready 3D scenes} constructed by OmniX, ready for free exploration, PBR-based relighting, and physical simulation.}
\label{fig:scene_application}
\end{figure}

OmniX enables automatic production of PBR-ready 3D scenes from images or panoramas. To evaluate the practicality of these generated 3D scenes, we import them into Blender and implement various graphics workflows, including free exploration, PBR-based relighting, and physical simulation, as shown in Figure~\ref{fig:scene_application}. Specifically, for free exploration, we move the camera forward to render a novel panoramic view. For {PBR-based relighting}, we add a point light source and animate its horizontal movement in a circular path around the scene’s center. For {physical simulation}, we introduce an elastic ball into the scene, assigning it an initial horizontal velocity to enable dynamic interactions within the environment. Demonstration videos are provided in the supplementary material.

%% file: tables/comp_appearance.tex
\begin{table}[t]
\centering
\caption{\textbf{Quantitative evaluation of OmniX on panorama generation} compared to existing image-to-panorama generation methods: OmniDreamer~\cite{omnidreamer}, PanoDiffusion~\cite{panodiffusion}, Diffusion360~\cite{diffusion360}, and CubeDiff~\cite{cubediff}.}
\label{tab:comp_appearance}
\setlength{\tabcolsep}{2pt}
\renewcommand{\arraystretch}{1.1}
\resizebox{\linewidth}{!}{
\begin{tabular}{lcccccccccc}
\toprule
\multirow{2}{*}{\textbf{Method}} & \multicolumn{5}{c}{\textbf{Laval Indoor}} & \multicolumn{5}{c}{\textbf{SUN360}} \\
 \cmidrule(lr){2-6}\cmidrule(lr){7-11}
& FID\,$\downarrow$ & KID ($\times 10^2$)\,$\downarrow$ & CLIP-FID\,$\downarrow$ & FAED\,$\downarrow$ & CS\,$\uparrow$ & FID\,$\downarrow$  & KID ($\times 10^2$)\,$\downarrow$  & CLIP-FID\,$\downarrow$ & FAED\,$\downarrow$ & CS\,$\uparrow$ \\
\hline
OmniDreamer & 71.0 & 5.17 & 23.9 & 19.2& -~~ & 92.3 & 8.89 & 51.7 & 30.4 & - \\
PanoDiffusion & 58.6 & 4.08 & 26.6 & 106.8 & -~~ & 52.9 & 3.51 & 28.9 & 98.0 & - \\
Diffusion360 & 33.1 & 2.07 & 16.9 & 23.7 & 26.38~~ & 45.4 & 3.73 & 18.5 & 12.6 & 22.89 \\
CubeDiff & \underline{9.5} & \textbf{0.32} & \underline{3.2} & \underline{18.4} & \underline{27.02}~~ & \textbf{25.5} & \underline{1.33} & \underline{8.1} & \textbf{7.6} & \underline{25.00} \\
OmniX (Ours) & \textbf{7.4} & \underline{0.41} & \textbf{2.5} & \textbf{5.2} & \textbf{28.47}~~ & \underline{39.4} & \textbf{0.66} & \textbf{6.5} & \underline{8.7} & \textbf{27.54} \\
\bottomrule
\end{tabular}
}
\end{table}

%% file: tables/comp_intrinsic.tex
\begin{table}[t]
\caption{\textbf{Quantitative results of OmniX on panoramic intrinsic estimation} compared to five competing methods: RGB$\leftrightarrow$X~\cite{rgbx}, MGNet~\cite{interiorverse}, IDArb~\cite{IDArb}, IID~\cite{intrinsic_image_diffusion}, and DiffusionRenderer~\cite{diffusion_renderer}.}
\label{tab:comp_intrinsic}
\begin{center}
\resizebox{\linewidth}{!}{
\setlength{\tabcolsep}{5pt}
\begin{tabular}{l cc cc cc cc}
\toprule
\multirow{3}{*}{\textbf{Method}} & \multicolumn{6}{c}{\textbf{PanoX-OutDomain}} & \multicolumn{2}{c}{\textbf{Structured3D}} \\
 & \multicolumn{2}{c}{\textbf{Albedo}} & \multicolumn{2}{c}{\textbf{Roughness}} & \multicolumn{2}{c}{\textbf{Metallic}} & \multicolumn{2}{c}{\textbf{Albedo}} \\
\cmidrule(lr){2-7} \cmidrule(lr){8-9}
& PSNR$\uparrow$ & LPIPS$\downarrow$ & PSNR$\uparrow$ & LPIPS$\downarrow$ & PSNR$\uparrow$ & LPIPS$\downarrow$ & PSNR$\uparrow$ & LPIPS$\downarrow$ \\
\midrule
RGB$\leftrightarrow$X & 6.35 & 0.591 & 8.18 & 0.628 & 4.38 & 0.720 & \underline{14.40} & \underline{0.350}\\
MGNet & 7.93 & 0.583 & 10.22 & 0.625 & 6.37 & 0.656 & 13.21 & 0.496\\
IDArb & 9.42 & 0.562 & 9.57 & 0.603 & 4.30 & 0.554 & 10.87 & 0.505 \\
IID & 10.25 & 0.640 & 10.09 & 0.631 & 7.89 & 0.726 & 10.80 & 0.539\\
DiffusionRenderer & \underline{10.91} & \underline{0.556} & \underline{10.45} & \underline{0.591} & \underline{14.45} & \underline{0.425} & {9.12} & {0.396} \\
OmniX (Ours) & \textbf{17.76} & \textbf{0.344} & \textbf{16.21} & \textbf{0.398} & \textbf{18.87} & \textbf{0.254} & \textbf{20.35} & \textbf{0.174} \\
\bottomrule
\end{tabular}
}
\end{center}
\end{table}

%% file: tables/comp_geometry.tex
\begin{table}[t]
\caption{\textbf{Quantitative results of OmniX on panoramic geometry estimation} compared to six state-of-the-art competing methods: DiffusionRenderer~\cite{diffusion_renderer}, MGNet~\cite{interiorverse}, DepthAnywhere~\cite{depthanywhere}, OmniData-v2~\cite{omnidata_v2}, DepthAnyCamera~\cite{depthanycamera}, and MoGe~\cite{moge}. For fair comparison, we use PanoX-OutDomain as the evaluation set to ensure all methods are evaluated in unseen scenarios. Note that MoGe uses far more depth annotations than ours (9.0M vs. 0.087M). Moreover, MoGe requires multi-view inference and stitching for panorama inputs, which is both complex and inefficient.}
\label{tab:comp_geometry}
\vspace{-18pt}
\begin{center}
\resizebox{\linewidth}{!}{
\setlength{\tabcolsep}{3pt}
\begin{tabular}{ll cccc cccc}
\toprule
\multirow{2}{*}{\textbf{Type}} & \multirow{2}{*}{\textbf{Method}} & \multicolumn{4}{c}{\textbf{Distance}} & \multicolumn{4}{c}{\textbf{Normal}} \\
\cmidrule(lr){3-6} \cmidrule(lr){7-10}
& & AbsRel$\downarrow$ & $\delta$-1.25$\uparrow$ & MAE$\downarrow$ & RMSE$\downarrow$ & Mean$\downarrow$ & Median$\downarrow$ & 5$^{\circ}$$\uparrow$ & 30$^{\circ}$$\uparrow$ \\
\midrule
\multirow{3}{*}{Depth-Only} & DepthAnywhere & 0.345 & 0.392 & 1.80 & 9.59 & / & / & / & / \\
& DepthAnyCamera & 0.199 & 0.680 & 1.93 & 7.86 & / & / & / & / \\
& MoGe & \textbf{0.106} & \textbf{0.898} & \textbf{1.04} & \textbf{5.35} & / & / & / & / \\
\midrule
\multirow{4}{*}{Geometry} & DiffusionRenderer & 0.709 & 0.246 & 2.55 & 16.10 & 97.19 & 89.62 & 0.001 & 0.023\\
& MGNet & 0.433 &  0.396 & 3.97 & 11.32 & \underline{79.96} & \underline{82.84} & 0.019 & \underline{0.269} \\
& OmniData-v2 & 0.342 & 0.440 & 1.94 & 10.76 & 85.22 & 100.60 & \underline{0.150} & 0.245 \\
& OmniX (Ours) & \underline{0.158} & \underline{0.787} & \underline{1.68} & \underline{6.83} & \textbf{27.14} & \textbf{14.88} & \textbf{0.155} & \textbf{0.663}\\
\bottomrule
\end{tabular}
}
\end{center}
\end{table}

%% file: tables/ablation_adapter.tex
\begin{table}[t]
\caption{\textbf{Ablation Studies.} We analyze the impact of 2D generative pre-trained weights and different adapter architectures on panoramic perception performance. Both datasets PanoX-Test and PanoX-OutDomain are used as the evaluation set to comprehensively cover both in-domain and out-domain scenarios.}
\label{tab:ablation_adapter}
\centering
\resizebox{\linewidth}{!}{
\setlength{\tabcolsep}{3pt}
\begin{tabular}{l cc cc cccc}
\toprule
\multirow{2}{*}{\textbf{Method}} & \multicolumn{2}{c}{\textbf{Albedo}} & \multicolumn{2}{c}{\textbf{Roughness}} & \multicolumn{4}{c}{\textbf{Distance}} \\
 \cmidrule(r){2-3} \cmidrule(r){4-5} \cmidrule(r){6-9}
& PSNR$\uparrow$ & LPIPS$\downarrow$ & PSNR$\uparrow$ & LPIPS$\downarrow$ & $\delta$-1.25$\uparrow$ & AbsRel$\downarrow$ & RMSE$\downarrow$ & MAE$\downarrow$ \\
\midrule
Shared-Branch     & 15.29 & 0.650 & 11.73 & 0.667 & 0.464 & 0.386 & 8.565 & 2.122 \\
Shared-Adapter    & 20.46 & 0.305 & 16.92 & 0.363 & 0.689 & 0.219 & 6.346 & 1.363 \\
Separate-Adapter (from scratch) & 17.74 & 0.534 & 14.83 & 0.510 & 0.382 & 0.461 & 8.771 & 2.729 \\
Separate-Adapter (Ours) & \textbf{21.68} & \textbf{0.260} & \textbf{18.16} & \textbf{0.329} & \textbf{0.808} & \textbf{0.154} & \textbf{4.755} & \textbf{1.110} \\
\bottomrule
\end{tabular}
}
\end{table}

%% file: sections/5_conclusion.tex
\section{Conclusion and Limitations}
\label{sec:conclusion}

In this work, we introduce OmniX, a versatile framework that repurposes pre-trained 2D flow-matching models for panorama generation, perception, and completion. We develop a unified formulation that casts both visual perception (RGB$\rightarrow$X) and visual completion (masked X$\rightarrow$X) into a 2D generative paradigm, and propose an efficient, lightweight cross-modal adapter to capture diverse task-specific knowledge. In addition, we construct a synthetic panorama dataset, PanoX, which spans indoor and outdoor environments and multiple visual modalities. PanoX serves as a comprehensive benchmark for panorama perception, addressing the scarcity of panorama data with dense geometric and material annotations. Extensive experiments demonstrate the effectiveness of our approach across panorama generation, perception, and completion tasks. Our approach further enables the automatic creation of immersive, PBR-ready 3D scenes that integrate seamlessly with standard 3D workflows.

\textbf{Limitations.}
Our method is built on top of pre-trained 2D flow matching models and thus inherits their shortcomings such as slow training and inference efficiency. In addition, OmniX's prediction of Euclidean distance is still not accurate enough, resulting in bumpy reconstructed 3D surfaces, which affects the subsequent PBR rendering effect. We also empirically observe that OmniX-Pano2Metallic, used for metallic prediction, performs poorly in generalization. This is partly due to the scarcity of panoramic PBR material data for training. Furthermore, the significant differences between neural rendering (\ie, 2D generative modeling) and PBR rendering may indicate that pre-trained 2D image priors have limited benefits for PBR material estimation.

%% file: sections/X_suppl.tex
\clearpage

\renewcommand{\thefigure}{S\arabic{figure}}
\renewcommand{\thetable}{S\arabic{table}}
\setcounter{figure}{0}
\setcounter{table}{0}

\section*{A. Additional Details of OmniX}
As a supplement to the unified formulation in Sec.~\textcolor{red}{3.3}, this section elaborates on how pre-trained 2D flow-matching models are repurposed for three panorama tasks: generation, perception, and completion.

Built upon the modality-specific LoRAs and circular synchronization operators, the OmniX framework can be flexibly applied to different panorama tasks and generate seamless results. Specifically, we consider three panorama task settings in this paper:

(\textbf{i}) For \textbf{panorama generation and completion}, we take a masked panorama as input condition and generate the complete panorama. This paradigm also supports image-to-panorama generation, where the masked panorama is defined as an empty panorama with the single-view input image projected onto it.

(\textbf{ii}) For \textbf{panorama perception}, \ie, RGB$\rightarrow$X, we take an RGB reference as the input condition and generate the target visual modality, such as distance, normal, albedo, roughness, and metallic. Optionally, additional conditions can be provided to improve performance, such as a camera ray map.

(\textbf{iii}) For \textbf{guided panorama perception}, we take both an RGB reference and a masked target as input conditions and generate the complete target modality. This setting is useful for progressive completion when building 3D scenes.

Among the above tasks, only panorama generation requires text prompts; the text prompt is set to an empty string for the other tasks.

%%%%%%%%%%%%%%%%%%%%%%%%%%%%%%%%%%%%%%%%%%%%%%%%%%%%%%%%%%%%
\section*{B. Results on Panorama Completion}
The panorama completion task involves inpainting panoramic maps of RGB and other modalities. We present both quantitative and qualitative results to evaluate the performance of OmniX-Fill.

\textbf{Datasets.} The data sources used for panorama completion are the same as those used for panorama generation and perception, except that we adopt the occlusion-aware data construction pipeline proposed in Sec. \textcolor{red}{3.4} to construct masked panoramas. These masked data, along with the corresponding masks and ground truths, are used for training and evaluation of OmniX-Fill.

\input{tables/comp_fill}

\textbf{Evaluation metrics.} We adopt the same evaluation metrics as in panorama generation to measure the similarity between the inpainted panoramas and the corresponding ground truths, including: FID~\cite{fid}, KID~\cite{kid}, CLIP-FID~\cite{clip_fid}, and FAED~\cite{panfusion}. FID, KID, and CLIP-FID are computed on perspective crops from generated ERP panoramas, while FAED is evaluated directly on ERP outputs.

\textbf{Quantitative evaluation.} We provide the quantitative results of OmniX-Fill on panorama completion compared to the state-of-the-art image inpainting method Flux-Fill, as reported in Table~\ref{tab:comp_fill}. OmniX-Fill consistently outperforms Flux-Fill across all metrics, demonstrating its superior performance on panorama completion under viewpoint changes and occlusion-induced masking, making it well-suited for iterative scene completion.

\begin{figure}[tbp]
\centering
\includegraphics[width=\linewidth]{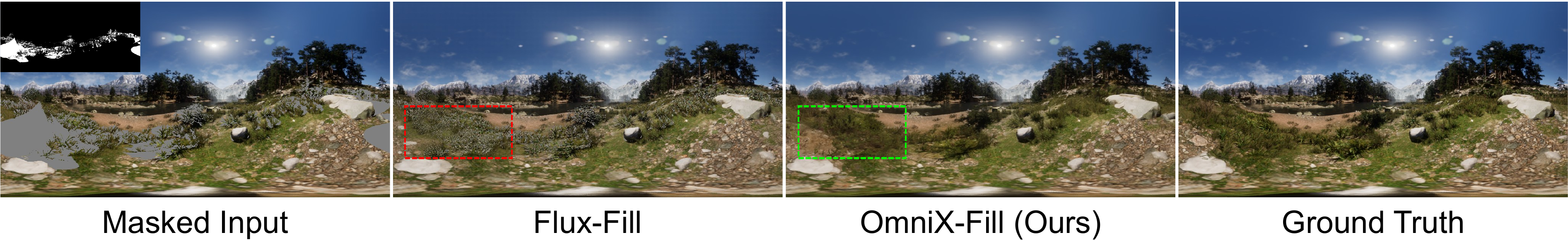}
\caption{\textbf{Qualitative evaluation of OmniX on panorama completion} compared to the state-of-the-art image inpainting method: Flux.1-Fill~\cite{flux_fill}.}
\label{fig:comp_fill}
\end{figure}

\begin{figure}[htbp]
\centering
\includegraphics[width=\linewidth]{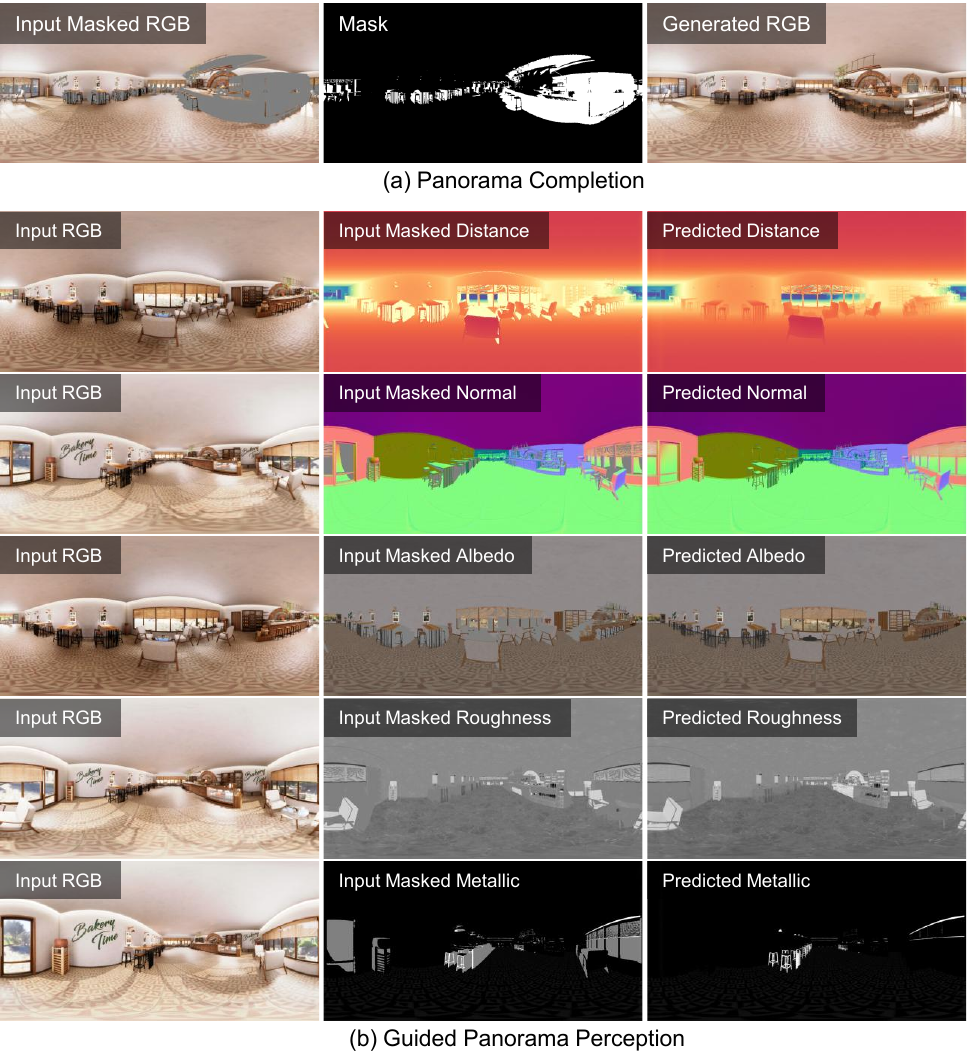}
\caption{\textbf{Panorama completion and guided panorama perception results} of OmniX-Fill. Given masked inputs, OmniX-Fill is able to generate accurate and locally coherent results for masked areas. For guided panorama perception, RGB references are necessary to ensure that all predicted modalities are aligned on the context.}
\label{fig:demo_pano_comp}
\end{figure}

\textbf{Qualitative evaluation.} Figure~\ref{fig:comp_fill} presents a qualitative comparison of panorama completion results between OmniX-Fill and the state-of-the-art Flux-Fill. The masked input shows regions missing due to viewpoint variations and occlusions, highlighting the challenges of realistic scene completion. Flux-Fill struggles to reconstruct fine structures within the masked areas, often producing artifacts and inconsistencies. In contrast, OmniX-Fill generates visually plausible and structurally consistent content, successfully preserving fine details such as vegetation and terrain contours. Additional results for panorama completion and guided panorama perception using OmniX-Fill are presented in Figure~\ref{fig:demo_pano_comp}, demonstrating that our method robustly handles complex masking scenarios and is ready for iterative scene completion.

%%%%%%%%%%%%%%%%%%%%%%%%%%%%%%%%%%%%%%%%%%%%%%%%%%%%%%%%%%%%
\section*{C. Additional Results on Panorama Perception}
\begin{figure}[htbp]
\centering
\includegraphics[width=\linewidth]{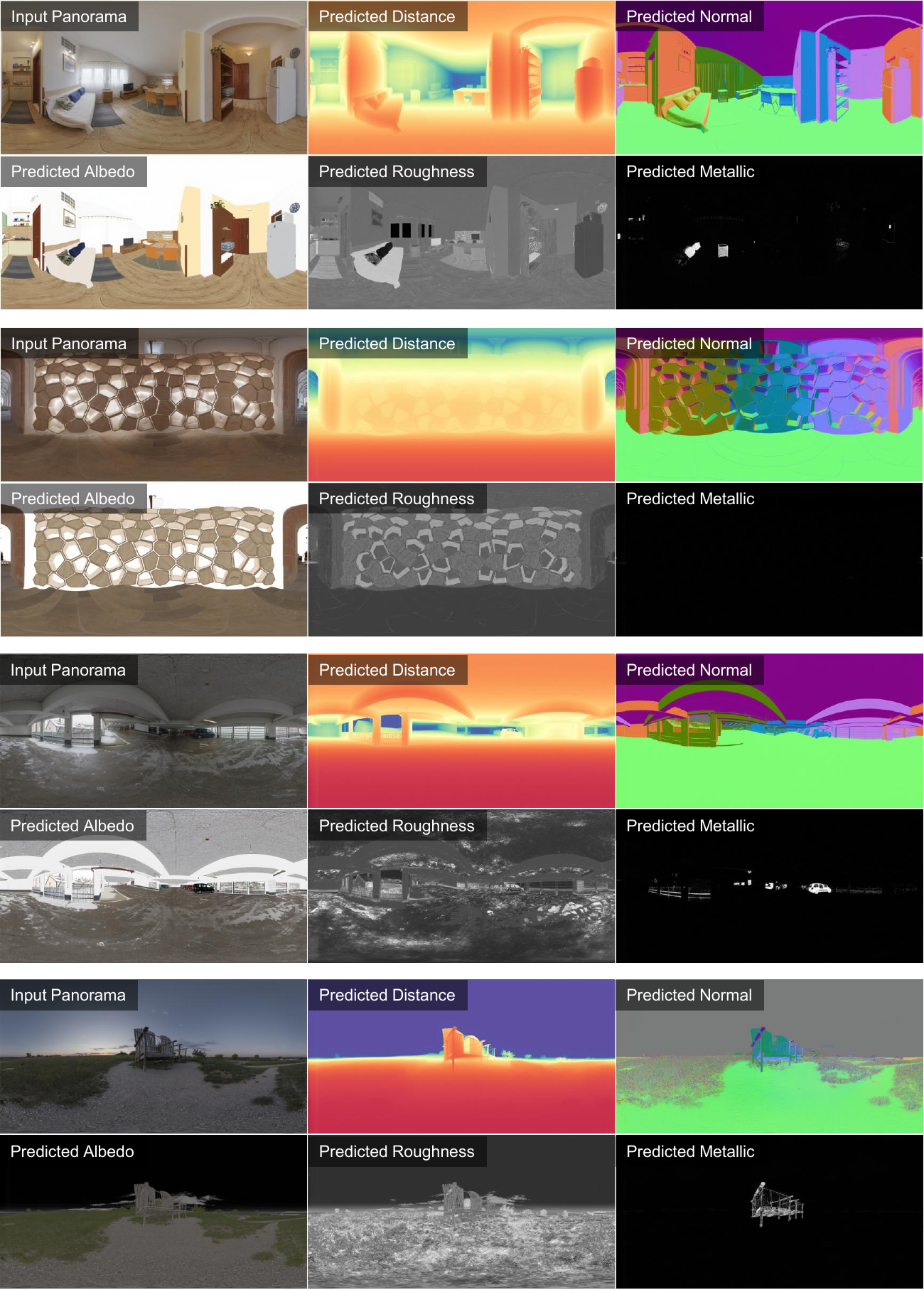}
\caption{\textbf{Additional panorama perception results} of OmniX on in-the-wild panoramas from the Internet. Our method exhibits strong generalization to unseen panoramas, thanks to the effective modality-specific adapter design that leverages pre-trained 2D generative priors for panorama perception.}
\label{fig:demo_pano_perc}
\end{figure}

We present additional panorama perception results of OmniX on in-the-wild images from the Internet in Figure~\ref{fig:demo_pano_perc}, highlighting the method’s strong generalization to unseen panoramas. This performance stems from our modality-specific adapter design, which effectively leverages the image generative priors of pre-trained 2D flow matching models for panorama perception.

%%%%%%%%%%%%%%%%%%%%%%%%%%%%%%%%%%%%%%%%%%%%%%%%%%%%%%%%%%%%

\begin{figure*}[htbp]
\centering
\includegraphics[width=\linewidth]{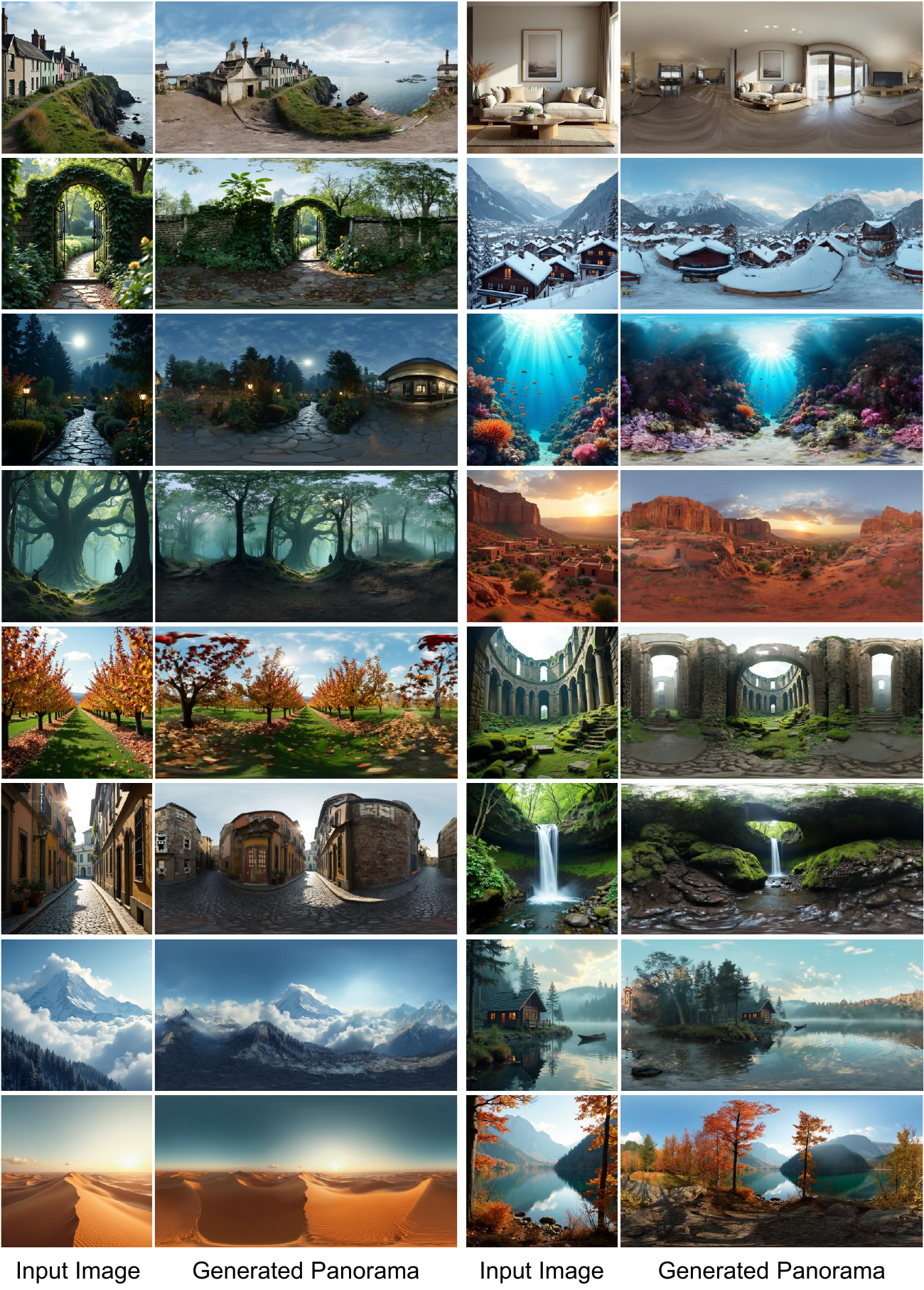}
\caption{\textbf{Additional panorama generation results} of OmniX given single-view image inputs generated by Flux.1-dev~\cite{flux1dev}. Our method produces photorealistic and diverse full-view panoramas.}
\label{fig:demo_pano_gen}
\end{figure*}

%%%%%%%%%%%%%%%%%%%%%%%%%%%%%%%%%%%%%%%%%%%%%%%%%%%%%%%%%%%%
\section*{D. Additional Results on Panorama Generation}
We present additional panorama generation results of OmniX from single-view image inputs in Figure~\ref{fig:demo_pano_gen}, demonstrating the method’s ability to produce high-quality and diverse panoramas. These results highlight OmniX’s capacity to infer missing spatial content and generate visually coherent details across the full 360° field of view, even from limited single-view observations.

%%%%%%%%%%%%%%%%%%%%%%%%%%%%%%%%%%%%%%%%%%%%%%%%%%%%%%%%%%%%
\section*{E. Additional Ablation Analysis and Discussion}
This section presents extended ablation studies and detailed discussions of the remaining components in our framework.

\textbf{Impact of camera rays as conditions.} Camera rays play an important role in spatial perception and scene understanding. We evaluate the effect of incorporating camera rays as an additional conditional input on panorama perception. As shown in Table~\ref{tab:ablation_camray}, including camera rays slightly improves the accuracy of normal map predictions, while yielding negligible gains for other modalities.

\input{tables/ablation_camray}

\textbf{Impact of different PBR material input arrangements.} VAEs used in 2D latent flow-matching models are trained on three-channel RGB inputs, preventing single-channel PBR material maps from being directly processed. Existing methods~\cite{intrinsic_image_diffusion,intrinsix} address this by concatenating roughness and metallic maps along with an additional zero channel to form a three-channel input. We investigate the effect of different PBR material input arrangements on visual perception performance, as summarized in Table~\ref{tab:ablation_joint_material}. Directly concatenating PBR material maps along the channel dimension is found to be suboptimal, leading to blurred and less accurate predictions. In contrast, jointly modeling PBR materials via a cross-attention mechanism significantly improves the results.

\input{tables/ablation_joint_material}

\input{tables/ablation_joint_geometry}

\textbf{Impact of joint distance-normal modeling.} Euclidean distance maps and normal maps are strongly correlated, so intuitively modeling them jointly should lead to improved performance. However, as shown in Table~\ref{tab:ablation_joint_geometry}, such joint geometry modeling does not bring positive performance gains for the prediction of either modality. This may be because the model fails to learn the geometric relationship between distance and normal vectors from the limited training data.

%% file: tables/comp_fill.tex
\begin{table}[htbp]
\centering
\caption{\textbf{Quantitative evaluation of OmniX-Fill on panorama completion} compared to the state-of-the-art image inpainting method: Flux.1-Fill~\cite{flux_fill}.}
\label{tab:comp_fill}
\resizebox{0.75\linewidth}{!}{
\setlength{\tabcolsep}{5pt}
\begin{tabular}{ l cccc }
\toprule
\multirow{2}{*}{\textbf{Method}} & \multicolumn{4}{c}{\textbf{PanoX-OutDomain (RGB)}} \\
\cmidrule(r){2-5}
& FID$\downarrow$ & KID($\times10^2$)$\downarrow$ & CLIP-FID$\downarrow$ & FEAD$\downarrow$ \\
\midrule
Flux-Fill & {30.09} & {0.975} & {5.74} & 2.53 \\
OmniX-Fill & \textbf{16.14} & \textbf{0.211} & \textbf{2.41} & \textbf{1.49}\\
\bottomrule
\end{tabular}
}
\end{table}

%% file: tables/ablation_camray.tex
\begin{table}[htbp]
\centering
\vspace{-5pt}
\caption{\textbf{Impact of camera raymaps as additional condition inputs.} We use PanoX-Test and PanoX-OutDomain jointly as the evaluation set to comprehensively cover both in-domain and out-of-domain scenarios.}
\label{tab:ablation_camray}
\setlength{\tabcolsep}{4pt}
\resizebox{\linewidth}{!}{
\begin{tabular}{l cc cc cc c c}
\toprule
\multirow{2}{*}{\textbf{Method}} & \multicolumn{2}{c}{\textbf{Distance}} & \multicolumn{2}{c}{\textbf{Normal}} & \multicolumn{2}{c}{\textbf{Albedo}} & \multicolumn{1}{c}{\textbf{Roughness}} & \multicolumn{1}{c}{\textbf{Metallic}} \\
 \cmidrule(r){2-3} \cmidrule(r){4-5} \cmidrule(r){6-7} \cmidrule(r){8-8} \cmidrule(r){9-9}
& AbsRel$\downarrow$ & $\delta$-1.25$\uparrow$ & Mean$\downarrow$ & Median$\downarrow$ & PSNR$\uparrow$ & LPIPS$\downarrow$ & PSNR$\uparrow$ & PSNR$\uparrow$ \\
\midrule
w/o CamRay  & \textbf{0.154} & \textbf{0.808} & 20.58 & 11.72 & \textbf{21.68} & \textbf{0.260} & \textbf{18.16} & 24.64 \\
w/ CamRay   & 0.155 & \textbf{0.808} & \textbf{19.92} & \textbf{10.99} & 21.29 & \textbf{0.260} & 17.59 & \textbf{25.52} \\
\bottomrule
\end{tabular}
}
\vspace{-5pt}
\end{table}

%% file: tables/ablation_joint_material.tex
\begin{table}[htbp]
\vspace{-5pt}
\caption{\textbf{Impact of different PBR material input arrangements.} We use PanoX-Test and PanoX-OutDomain jointly as the evaluation set to comprehensively cover both in-domain and out-of-domain scenarios.}
\label{tab:ablation_joint_material}
\centering
\resizebox{0.87\linewidth}{!}{
\setlength{\tabcolsep}{6pt}
\begin{tabular}{l cc cc cc}
\toprule
\multirow{2}{*}{\textbf{Method}} & \multicolumn{2}{c}{\textbf{Roughness}} & \multicolumn{2}{c}{\textbf{Metallic}} & \multicolumn{2}{c}{\textbf{Average}} \\
\cmidrule(r){2-3} \cmidrule(r){4-5} \cmidrule(r){6-7}
& PSNR$\uparrow$ & LPIPS$\downarrow$ & PSNR$\uparrow$ & LPIPS$\downarrow$ & PSNR$\uparrow$ & LPIPS$\downarrow$ \\
\midrule
joint (concat.)       & 17.66 & 0.350 & 24.58 & 0.323 & 21.12 & 0.337 \\
joint (cross-attn.)   & 17.43 & 0.340 & \textbf{25.43} & \textbf{0.138} & \textbf{21.43} & \textbf{0.239} \\
independent           & \textbf{18.16} & \textbf{0.329} & 24.64 & 0.153 & 21.40 & 0.241 \\
\bottomrule
\end{tabular}
}
\end{table}

%% file: tables/ablation_joint_geometry.tex
\begin{table}[t]
\caption{\textbf{Impact of joint distance-normal modeling.} We use PanoX-Test and PanoX-OutDomain jointly as the evaluation set to comprehensively cover both in-domain and out-of-domain scenarios.}
\label{tab:ablation_joint_geometry}
\centering
\setlength{\tabcolsep}{4pt}
\resizebox{\linewidth}{!}{
\begin{tabular}{l cccc cccc}
\toprule
\multirow{2}{*}{\textbf{Method}} & \multicolumn{4}{c}{\textbf{Distance}} & \multicolumn{4}{c}{\textbf{Normal}} \\
\cmidrule(r){2-5} \cmidrule(r){6-9}
& AbsRel$\downarrow$ & $\delta$-1.25$\uparrow$ & MAE$\downarrow$ & RMSE$\downarrow$ & Mean$\downarrow$ & Median$\downarrow$ & 5$^{\circ}$$\uparrow$ & 30$^{\circ}$$\uparrow$\\
\midrule
joint  & 0.163 & 0.787 & 1.11 & 5.35 & 20.80 & 11.95 & 0.227 & 0.767 \\
independent          & \textbf{0.155} & \textbf{0.808} & \textbf{1.08} & \textbf{5.35} & \textbf{19.92} & \textbf{10.99} & \textbf{0.249} & \textbf{0.779} \\
\bottomrule
\end{tabular}
}
\vspace{-5pt}
\end{table}